# Framework for Hopfield Network based Adaptive routing

A design level approach for adaptive routing phenomena with Artificial Neural Network

By
**Shankar.R.**, M.S. Research, student, Department of Electronics and Communication, Anna University, Chennai, India

Under the guidance of
**Prof Dr.N.Gunasekaran.**, Ph.D.,
Head of the Department, Department of Electronics and Communication, Anna University, Chennai, India

**Introduction :**
Routing, as a basic phenomena, by itself, has got umpteen scopes to analyse, discuss and arrive at an optimal solution for the technocrats over years. Routing is analysed based on many factors; few key constraints that decide the factors are communication medium, time dependency, information source nature. Parametric routing has become the requirement of the day, with some kind of adaptation to the underlying network environment.
Satellite constellations, particularly LEO satellite constellations have become a reality in operational to have a non-breaking voice/data communication around the world. (Ref : Iridium - operational : www.iridium.com)

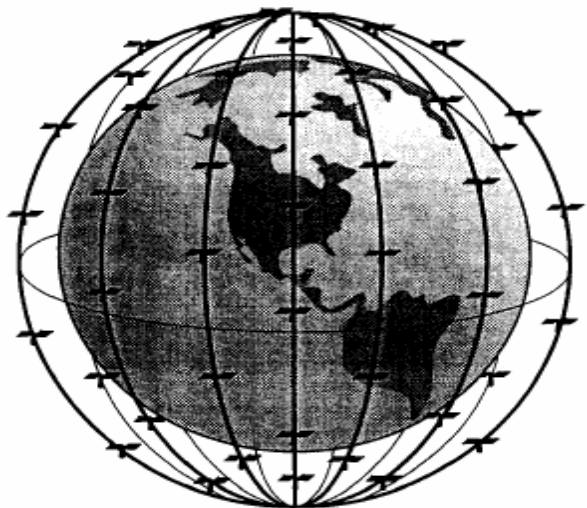

**77 Low Earth Orbit Space Vehicles**
- Polar orbit at 765 kilometers
- Full earth coverage

Routing in these constellations has to be treated in a non conventional way, taking their network geometry into consideration.
One of the efficient methods of optimization is putting Neural Networks to use. Few Artificial Neural Network models are very much suitable for the adaptive control mechanism, by their nature of network arrangement. One such efficient model is Hopfield Network model.

This paper is an attempt to design a framework for the Hopfield Network based adaptive routing phenomena in satellite constellations and is arranged into the following parts. Part 1 discusses about the LEO satellite constellation in general and it's communication details. Part 2 discusses about artificial neural networks in general and Hopfield in particular. Part 3 discusses about the approach to routing optimization by TSP optimization with Hopfield. Part 4 discusses about the framework and design parameters of the concerned subject. Part 5 supporting Matlab code and graphs.

**Part 1 : LEO satellite constellation**

LEO orbit is used by satellite constellations for personal communication services because of advantages in terms of low launch cost, low time delay for signal transfer. Orbital height – prime constraint which decides life time, regional propagation time delay, space craft weight and hence the initial system launch cost
Prime reasons for deploying LEO constellation
1. Increased network survivability , redundancy
2. Coverage of entire earth especially polar regions
3. Reduction of much wanted geostationary orbit congestion.

Commercial communication services takes place in L band – 20 GHz. Opportunistic customers are international (transatlantic, transpacific) & national roamers, military agencies for remote hilly area coverage. LEO satellite communication requires a low link margin – approx. 16 db – this permits usage of hand held units even in closed environment. LEO orbits are usually circular orbit with x degree inclination. LEO constellation requires tightly focused beams. Ground to space Communication via SCPC / TDMA & Space to ground communication, ISL are via continuous TDM. Satellites behave like tandem switches in LEO Constellation.
Service area of a satellite is called its foot print, which is a spherical segment of the earth's space in which the satellite can be seen.
Many numbers of satellites are required to cover the entire earth to give a complete coverage. The number of satellites and number of orbits depends upon the constellation geometry. Inter-satellite links are used to carry traffic between the satellites to have a continuous communication; in case of satellite movement causing loss of visibility or user movement causing loss of visibility. The LEO satellite communication geometry is arranged in such a way that the satellite's coverage is marked by logical position. Such logical positions remain static even if the satellites move; causing the next satellite to cover the logical position of the previous satellite. This creates complication in communication hand-off in-turn in the call routing.

As LEO satellite constellation is primarily used for personal communication services like voice and data, much of routing have to be done for each communication service call. The constellation's physical geometry by itself has many obstacles in terms of network non reach-ability and seam between the orbits of the constellation. The routing should also address the service category in terms of time criticalness, bandwidth requirement in terms of constant bit rate requirement, variable bit rate requirement and available bit rate requirement applications. Statistical models of the source traffic activity have to be incorporated. The traffic has to be modelled as stochastic processes – family of random variables which are functions of time given as closed form probability distribution.

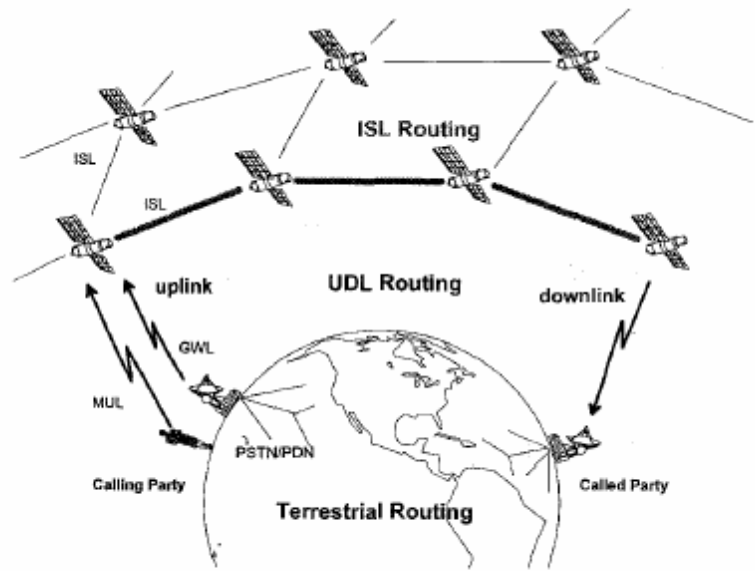

Fig. 1. Routing scenario in a LEO system with ISLs.

The LEO ISL environment has varying time and spatial variance of traffic flows. Moreover the personal communication requires high QoS which demands efficient adaptive routing schemes in order to achieve optimal network performance. These add additional burden to the conventional routing complexities.

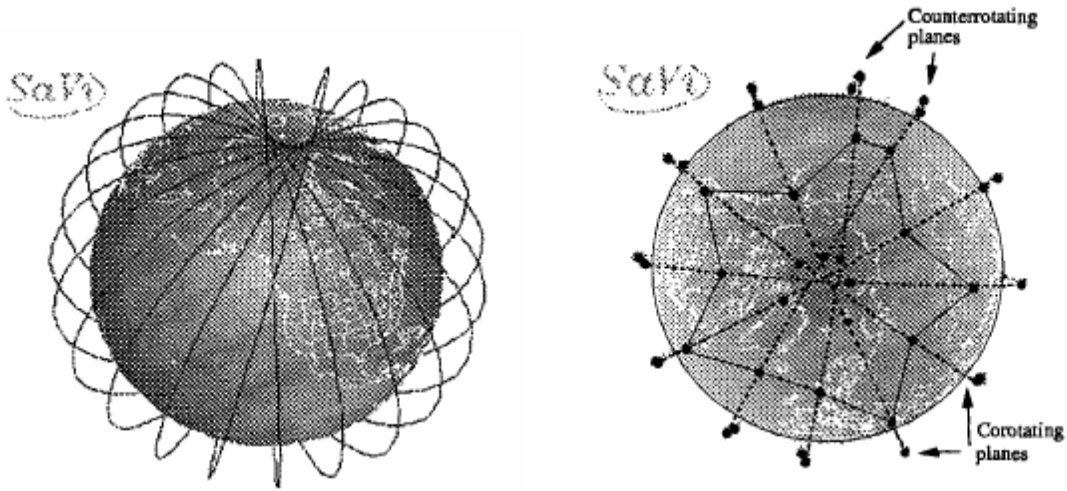
Figure explaining the seam factor in satellite constellation.

**Part 2 : Artificial Neural Network – Hopfield Network**
Artificial Neural Networks are biologically inspired computing models characterized by large number of densely connected computational elements (neurons). The inherent characteristics of the NN i.e. the parallel architecture and individual computing neurons make artificial neural networks obvious choice for solving optimization problems. It is basically motivated by the idea that due to high complexity of the problem it may show a superior performance compared to deterministic algorithm approaches. Moreover, since ANN is based on "hidden intelligence" gained in one or several learning phase(s) – rather than on permanent information exchange throughout the network; it inherently avoids the need for large signalling overhead and related delay problems. Hopfield neural network model is a fully optimization network of simple processing units with numerically weighted symmetric connections. Hopfield network has advanced prominence because of the relative ease of building it into hardware using currently available VLSI technology. Each processing element performs a simple and identical computation which generally involves summing weighted inputs to the unit, applying the transfer function and changes the state if necessary. The power of the Hopfield network lies in the connection between the units and the weights of the connections.
The Hopfield defines a energy function of it's own called Liapunov function. The Liapunov function is bounded from below, that is from any starting state the network would always converge to some energy function minimum, upon applying a sequence of asynchronous local state updates (that locally reduced energy). The parameters of the energy function should be chosen in a way to reduce the necessary values to achieve optimum result.

**Part 3 : TSP approach to Routing problems via Neural Network**
The functionality of the routers is to route the data packets (or switch to receiver in case of voice) to the receiving end. It is simply route selection for the incoming origin packet for every origin destination pair(s), delivery of the packets to the corresponding destination point for every origin data packet. This is achieved with the help of routing tables. The routing process is distributed; routing tables decide the next hop, which could be a next router or the destination itself. To achieve the distributed routing little information exchange is made between the nodes and routing nodes by itself learn and adapt to the network environment.

The functionality of the above can be compared with that of a travelling sales man problem. In TSP, the obvious constraints like non availability at more than one place, one time and minimum distance travelled are also similar to the routing packets. The constraints can be customized for the routing problem and could be incorporated with varying weights.

Few constraints with respect to our problem would be :
- Not more than one node shall have control of routing packet
- Not more than time a node shall be visited by a packet
- The distance travelled by the packet should be minimum

These constraints can be incorporated into the Hopfield network by virtue of its design weight-age. Also a network distance matrix would help in assigning the initial weights to the connections. The optimization is sensitive to fine tuning of the coefficients of the Liapunov energy function, which directly maps of the problem constraints. Linear threshold activation is used in the transfer function. The energy minimization happens with the state updates. The training patterns will become stable attractors. There could be some local minima that would be spurious attractors; that have to be overcome with repeated training and number of loop operations.

**Part 4 : Framework and design parameters**

A Hopfield network based adaptive distributed routing solution for the LEO satellite constellation would not be a mere shortest path routing solution. The routing solution to be obtained has to take care of the below key issues

1. ISL vs non ISL based connectivity
2. Seam in orbital planes for closed earth coverage
3. Nature of voice / data circuit connectivity – real time vs non real time
4. Circuit handoff with respect to satellite movement and user movement
5. Evenly balancing network load – because global call traffic is not uniform and geography, local time dependant

Addressing the above issues, all operational LEO satellite constellations are provisioned with ISLs with connectivity routing among them. ISLs could be treated as alternate paths with viable routes to the destination, but might be with relatively longer network delay.

The seam in the orbital planes makes the cross inter-satellite link non functional between adjoint orbital planes, because of the opposite direction movement of satellites and non availability of pointing controls for the satellite antenna in cross seam. This seam makes the nearby satellites to communicate via a relatively long distance; because the packet travels till the one of the poles and comes back. Also this makes the routing complex and a requirement of arriving at a connectivity matrix to denote the long distance path. This matrix aids in arriving at the weights of the Hopfield network for path computation.

The nature of data/voice circuit decides the requirement of network parameters viz signal latency, round-trip-delay, network uptime and availability, handoff signalling mechanism, alternate routes availability, communication gateway. Signal latency is of importance for a voice only network, round trip delay would be of more important for constant-bit-rate applications like live video / audio broadcast, network uptime is a overall quality determining factor and availability is critical for real time applications, Circuit handoff for the voice call to have a non-break service is critical which should be facilitated by a separate handoff mechanism and also should be supported to routing systems, logical locations of satellite should be used for this kind of soft-handover mechanisms. The availability of the alternate routes and the routing intelligence to update itself is the key issue for redirection of data packets in case of congestion in a single favoured short path. The communication gateway availability for the satellite based traffic to merge into the terrestrial traffic. Adequate number of gateways with routing intelligence to select the destination gateway is a key issue in service deliverables.

All the above could be incorporated in the weight matrices and/or coefficients of energy function.

Implementing the above in a test run format is the idea behind my masters in science program.

**Part 5 : Programming the concepts**

Programming the Hopfield network in Matlab environment has it's own advantages in terms of data manipulation and date interpretation.

Matlab with Neural Network toolbox provides avenues coding with multiple levels of iteration.

The coding developed is based on the constraints mentioned in part 3 of this article.

This code is developed with following assumptions and reality randomness
1. No of nodes : 6
2. No of hops permitted : 6
3. Distances across nodes are known
   (Distances are directly proportional to switching and packet travel time)
4. Bandwidth efficiency of channels has normal distribution with mean occupancy of 50 % and variance 25 % total in both directions
5. Weights are calculated with real time reality conditions
6. Hopfield neural network is utilised to find optimum route and decreasing Liapunov energy function is confirmed with negative energy growth in iterations amidst local minimas.
7. Result gives packet traversal path with arbitrary data origination and constrained hopping.

**Coding :**

**Initnetwork.m**

```
nNodes = 6;
threshold = 31;
nHops = nNodes;
nInputs = nNodes * nHops;
inhibWeight = -3;
offState = -1; %%% Binary threshold function should return a 1 or a -1.
inputWidth = nNodes;
initialActivations = ones(nInputs,1) * -1;
activations = initialActivations;

%%% Assign x,y coordinates to each city; then calculate distances.

nodeLocations = zeros(nNodes,2);
nodeLocations(1,:) = [0 3];
nodeLocations(2,:) = [2 5];
nodeLocations(3,:) = [5 3];
nodeLocations(4,:) = [3 2];
nodeLocations(5,:) = [4 0];
nodeLocations(6,:) = [2 1];

throughput=zeros(6,6)
throughput=normrnd(.5,.25,6,6)

%%%% Calculate distances between all pairs of cities.

distances = zeros(nNodes,nNodes);
for node1 = 1:nNodes,
  for node2 = 1:nNodes,
    difference = nodeLocations(node1,:) - nodeLocations(node2,:);
    distances(node1,node2) = sqrt(difference(1,1)  * difference(1,1) + difference(1,2)  * difference(1,2));
  end
end
%% Normalize distances to 1.0;
distances=distances*throughput

distances = distances / max(max(abs(distances)));
%%% Calculate weights;

weights = zeros(nInputs,nInputs);
for node1 = 1:nNodes,
  for hop1 = 1:nHops,
```

```matlab
    for node2 = 1:nNodes,
      for hop2 = 1:nHops,
        %%% Constraint I: For a valid routing, the same Node need
        %%% not appear at two different hops.
        %%% Make large inhibitory weight.
        if ((node1 == node2) & (hop1 ~= hop2))
           weights = setWeight(node1,hop1,node2,hop2,inhibWeight,weights,nNodes);
        %%% Constraint II: For a valid routing, there should not
        %%% be two different nodes at the same hop on the network.
        %%% Make large inhibitory weight.
        elseif ((hop1 == hop2) & (node1 ~= node2))
            weights = setWeight(node1,hop1,node2,hop2,inhibWeight,weights,nNodes);
        %%% Constraint III: For an optimal path routing, nodes at
        %%% adjacent hops on the tour should be short distance
        %%% apart. Make weight equal to negative of the distance.
        elseif ((node1 ~= node2) & (hop1 ~= hop2)) & (abs(hop1 - hop2) == 1)
           weights = setWeight(node1,hop1,node2,hop2,-distances(node1,node2),weights,nNodes);
        end
      end
    end
  end
end
clc
disp('parameters updated')
```

**setweight.m**

```matlab
%% weights is a matrix with each row representing the
%% incoming weights of a different unit.
%% The rows and columns in the weight matrix correspond to input/output units
%%  in the following order:
%% node1-hop1, node2-hop1, ... , nodeN-hop1, node1-hop2, node2-hop2,...,
%%% nodeN-hop2,  ... , node1-hopN, node2-hopN, ... , nodeN-hopN
%%
%% This function changes the value of a single weight in
%% the matrix to have the value 'wt'. It also sets the value of the
%% symmetric weight.
function [weights] = setWeight(node1,hop1,node2,hop2,wt,weights,nNodes)
  unit1 = (hop1 - 1) * nNodes + node1;
  unit2 = (hop2 - 1) * nNodes + node2;
  weights(unit1,unit2) = wt;
  weights(unit2,unit1) = wt;
```

### hopnetwork.m

```
%%% forward pass for hopfield program
k1=0;k=0;ksub=0;kgrand=0;count=1;kall=0;unit=0
energystore=zeros(36,1);
k1=0;k=0;ksub=0;kgrand=0;io=0;jo=0;qq=1;
activationsresult=zeros(36,36);
for alpha=1:36
order=randperm(nInputs);
order=order';
for i = 1:nInputs
   unit = order(i,1)
   oldActivation = activations(unit,1);
   netIn = weights(unit,:) * activations;
   if netIn > threshold
       activations(unit,1) = 1;
   else
     activations(unit,1) = offState;
   end
end

activationsresult(:,qq)=activations(:,1);
qq=qq+1;

k1=0
k1=weights(unit,:)*activations;
k=0
for io=1:36
   k=k1*activations(io,1);
   ksub=ksub+k
end
 ksub=-ksub/2;
kgrand=ksub

k1=0;k=0;ksub=0
for i=1:36
   k1=activations(i,1)*threshold
   k=k+k1
end

kall=kgrand+k
energystore(count,1)=kall
count=count+1;
end
```

```
min_energy=min(energystore)
toggle=0
count=0
while toggle==0
     count=count+1
   if min_energy==energystore(count)
    toggle=1
   end
end

energystr=int2str(energystore);

figure(1);
hold on
xlabel(' Iteration ')
ylabel('Energy level ')
plot(1:count,energystore(1:count),'b:',1:count,energystore(1:count),'ko')
for alpha = 1:count,
       text(alpha,energystore(alpha),energystr(alpha,:));
end
n = inputWidth;
activationGrid = zeros(n+1,n+1);
activations(:,1)=activationsresult(:,count)
activationGrid(1:n,1:n) = reshape(activations,n,n);

figure(2);
shading interp
pcolor(activationGrid);
xlabel('Hops')
ylabel('Nodes')
set(gca,'XTick',[1:nNodes]);
set(gca,'YTick',[1:nNodes]);

figure(3);
clf; hold on;
xlabel('Hops')
ylabel('Nodes')
maxX = max(nodeLocations(:,1)) + 1;
maxY = max(nodeLocations(:,2)) + 1;
minX = min(nodeLocations(:,1)) - 1;
minY = min(nodeLocations(:,2)) - 1;
plot(nodeLocations(:,1),nodeLocations(:,2),'o');
axis([minX maxX minY maxY]);
tour = zeros(nNodes * nNodes,2);
hopNum = 1;
```

```
for hop = 1:nNodes,
  for node = 1:nNodes,
    if activationGrid(node,hop) == 1
      tour(hopNum,:) = nodeLocations(node,:);
      hopNum = hopNum + 1;
    end
  end
end
numhops = hopNum -1;
alpha=0
for alpha=1:nNodes
   plot(tour(1:alpha,1),tour(1:alpha,2),'ks:');
end
end

for node = 1:nNodes,
  nodeName = sprintf('node %d',node);
  text(nodeLocations(node,1),nodeLocations(node,2),nodeName);
end
clc
disp('Network converged.')
```

**The random throughput value used was :**

| | | | | | |
|---|---|---|---|---|---|
| 0.38376 | 0.75946 | 0.98935 | 0.79756 | 0.5215 | 0.34218 |
| 0.59274 | 0.40255 | 0.62614 | 0.22095 | 0.00114 | 0.08130 |
| 0.68207 | 0.15468 | 0.96613 | 0.65882 | 0.37673 | 0.192 |
| 0.9028 | 0.57889 | 0.41505 | 0.34965 | 0.61551 | 0.76391 |
| 0.16068 | 0.88831 | 0.21506 | 0.6378 | 0.41975 | 0.47169 |
| 0.24435 | 0.67697 | 0.44722 | 0.22504 | 0.80914 | 0.59481 |

When this code is run with the given arbitrary values, we get the following graphs.

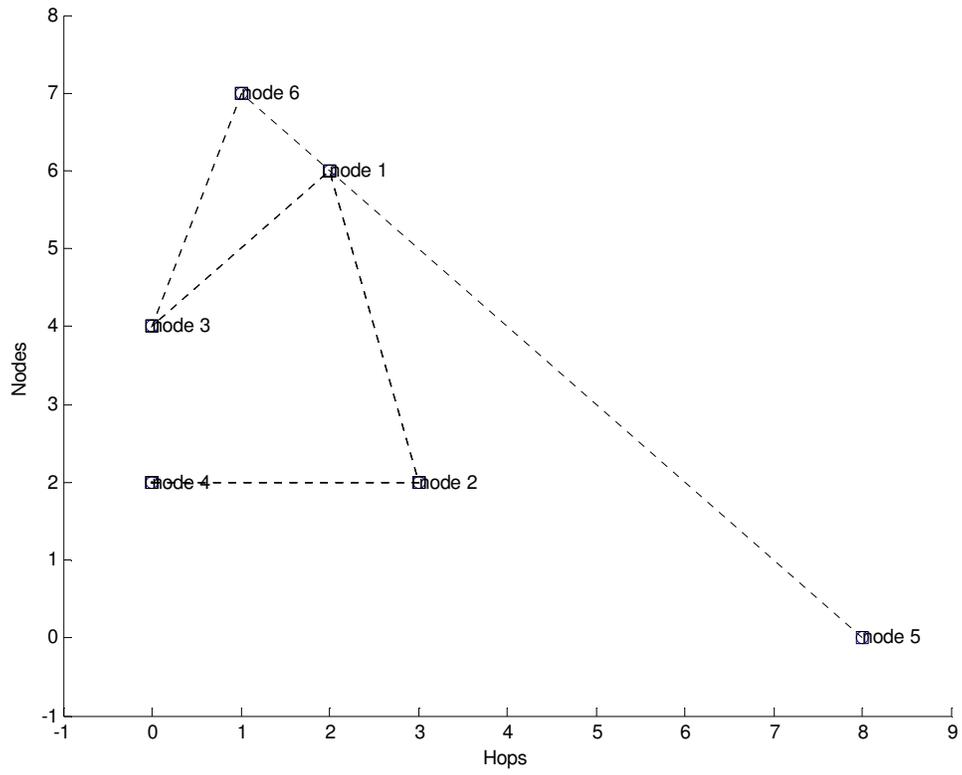

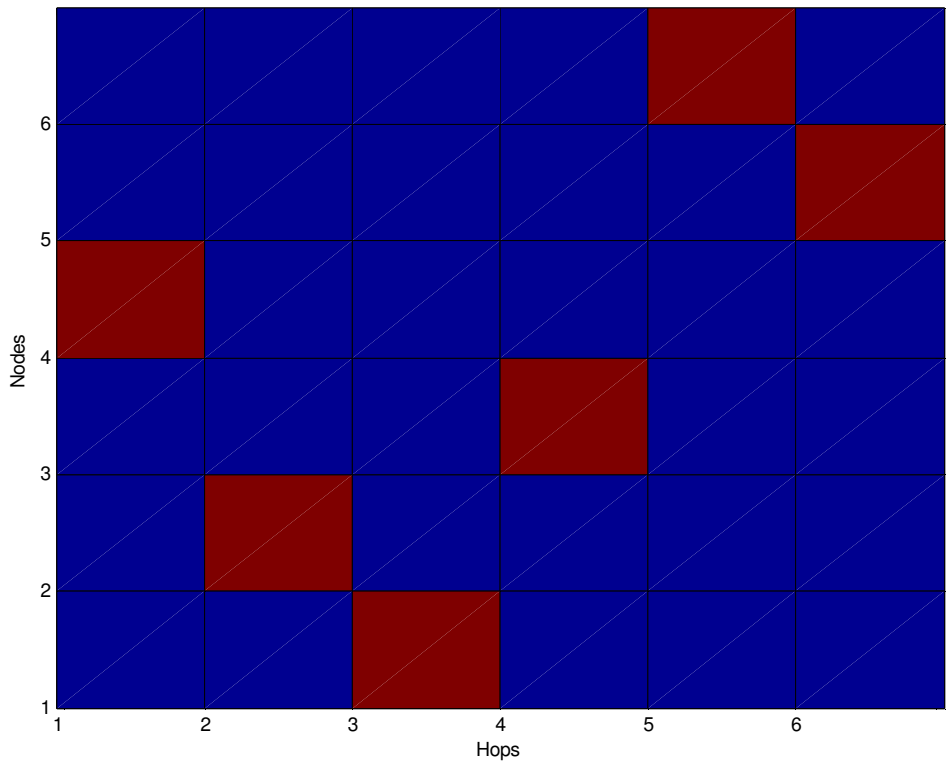

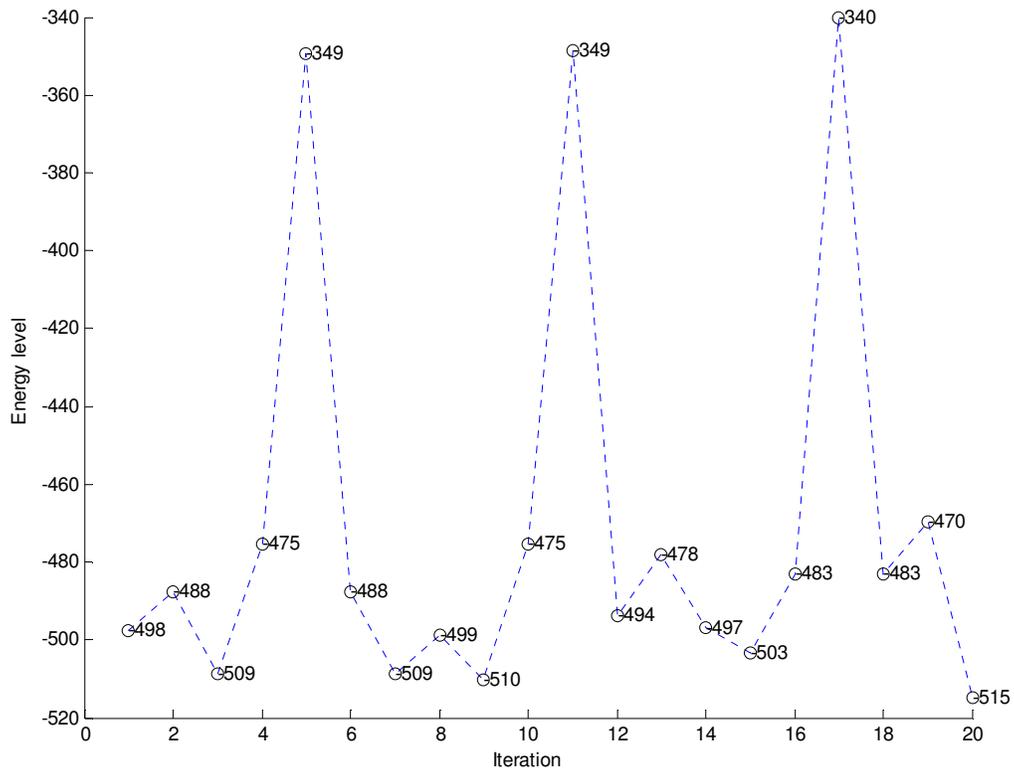

Conclusion :

Hopfield network shows it's superior performance over conventional methods of computing. The optimization time taken by the Hopfield network is relatively smaller and quality of optimization will also be better, in terms of energy consumption by the function. Also the current availability of VLSI chips implementing the Hopfield network is abundant and realising a program into reality becomes more of ease.
The LEO constellation is a long way to go in terms of technology maturity and Hopfield network based routing optimization is a step forward in that.